\documentclass[11pt,a4paper]{article}
\usepackage[hyperref]{naaclhlt2019}
\usepackage{times}
\usepackage{latexsym}

\usepackage{url}

\usepackage{graphicx}
\usepackage{amsmath}
\usepackage{booktabs} 

\aclfinalcopy 

\title{Improving Cross-Domain Chinese Word Segmentation with Word Embeddings}

\author{Yuxiao Ye$^1$\thanks{\ \ Yuxiao Ye and Yue Zhang contributed equally to this work.},  Yue Zhang$^2$\footnotemark[1]\ \thanks{\ \ Corresponding authors.}\ ,Weikang Li$^3$, Likun Qiu$^1$\footnotemark[2]\ , Jian Sun$^1$\\
  $^1$Alibaba Group, $^2$Peking University, $^3$Westlake University  \\
  {\tt \{yuxiao.yyx, likun.qlk, jian.sun\}@alibaba-inc.com}\\ 
  {\tt yue.zhang@wias.org.cn}\\
  {\tt wavejkd@pku.edu.cn} \\
  }

\date{}

\begin{document}
\maketitle
\begin{abstract}
Cross-domain Chinese Word Segmentation (CWS) remains a challenge despite recent progress in neural-based CWS. The limited amount of annotated data in the target domain has been the key obstacle to a satisfactory performance. In this paper, we propose a semi-supervised word-based approach to improving cross-domain CWS given a baseline segmenter. Particularly, our model only deploys word embeddings trained on raw text in the target domain, discarding complex hand-crafted features and domain-specific dictionaries. Innovative subsampling and negative sampling methods are proposed to derive word embeddings optimized for CWS.  We conduct experiments on five datasets in special domains, covering domains in novels, medicine, and patent. Results show that our model can obviously improve cross-domain CWS, especially in the segmentation of domain-specific noun entities. The word F-measure increases by over 3.0\% on four datasets, outperforming state-of-the-art semi-supervised and unsupervised cross-domain CWS approaches with a large margin. We make our code and data available on Github.
\end{abstract}

\section{Introduction}
Chinese Word Segmentation (CWS) is the first step for many Chinese Natural Language Processing (NLP) tasks \cite{cai2016neural,zhao2017hybrid}. Approaches to CWS could be categorized into two categories: character-based and word-based. The former treats CWS as a sequence labeling problem, labeling each character in a sequence with \emph{B/I/E/S} (\emph{Beginning, Internal, End, Single}) labels \cite{tseng2005conditional}. Traditional character-based approaches often use Conditional Random Fields (CRF) models to label sequences, with complex hand-crafted discrete features \cite{peng2004chinese,tseng2005conditional}. Unlike character based CWS, word-based CWS operates on a word-level, directly exploiting word-level features. Typical CRF models are replaced with semi-CRF models, in which labels are assigned to subsequences instead of characters \cite{sarawagi2005semi,liu2014domain}. Transition-based approaches have also been used to exploit larger feature contexts \cite{zhang2007chinese}. More recent approaches exploit neural networks including Recurrent Neural Networks (RNN) to replace hand-crafted discrete features with real-valued features \cite{cai2016neural,chen2015long,chen2017adversarial}.

Existing studies have achieved satisfactory results for in-domain CWS, with F-scores over 96.0\% in the newspaper domain \cite{chen2017adversarial}. Nevertheless, cross-domain CWS remains a big challenge \cite{liu2014domain,liu2012unsupervised}. The main reason is the lack of annotated data in the target domain, which makes supervised approaches less useful. To tackle this problem, some unsupervised and semi-supervised approaches have been proposed. One way is to exploit complex features including character types, lexical features and accessor varieties \cite{wu2014leveraging}, which requires much efforts on feature engineering. Another way is to deploy machine learning algorithms including self-training and model ensemble \cite{gao2010multi,liu2012unsupervised,qiu2015word}, which is time-consuming and inefficient. 

In this paper, we investigate a different approach to deploying unsupervised data for cross-domain CWS, in order to completely break free from the reliance on manual annotation, complex feature engineering, and even parametric training to some extent. We propose a Word-Embedding-Based CWS (WEB-CWS) model, which aims to improve the performance of an existing baseline segmenter in cross-domain CWS. WEB-CWS is a conceptually simple word-based model, using word embeddings, which are expected to carry semantic and syntax information \cite{mitchell2010composition}, as the only input of a non-parametric word segmentor. The basic intuition is that embeddings of words within a same context window should be close to each other \cite{goldberg2014word2vec,mikolov2013distributed}. If a sequence is incorrectly segmented, those incorrectly segmented words are likely to be semantically and syntactically inconsistent with their surrounding words. Consequently, the embedding of an incorrectly segmented word should be far away from embeddings of its surrounding words.

Based on the hypothesis above, we propose WEB-CWS. Word embeddings are first derived with a CWS-oriented word embedding model with innovative subsampling and negative sampling methods. A word-embedding-based decoder is then used for segmentation, with cosine similarities among word embeddings as the metric for probability calculation. WEB-CWS is a \emph{semi-supervised} model, because it only uses word embeddings trained on raw text in the target domain, which is first automatically segmented by the baseline segmenter. The model is also \emph{cross-domain} in the sense that it can improve the performance of the baseline segmenter, when the source text for training the baseline segmenter and the target text to be segmented are in different domains.

The main contributions of this paper include:

\begin{itemize}
\item To our knowledge, we are the first to directly use word embeddings for CWS, without any neural structures, which makes our model conceptually simpler and run faster.
\item We have proposed novel sampling methods to make the embeddings optimized for CWS, which has never been used for embedding training.
\item Our model can be used on top of any existing CWS models to improve their performances, without the need to re-train those models with annotated domain specific data.
\item On four datasets in different special domains, our model improves the word F-measure by more than 3.0\%, compared with the state-of-the-art baseline segmenter. We release our code and data on Github\footnote{\ \ https://github.com/vatile/CWS-NAACL2019}.
\end{itemize}

\section{Related Work}

Our work is related with existing research on word-based CWS, cross-domain CWS, and embedding-based CWS. 

\subsection{Word-Based CWS}
Instead of labeling a sequence character-wise, word-based CWS tries to pick the most probable segmentation of a sequence. \citeauthor{zhang2007chinese} \shortcite{zhang2007chinese} design a statistical method for word-based CWS, extracting word-level features directly from segmented text. The perceptron algorithm \cite{collins2002discriminative} is used for training and beam-search is used for decoding. \citeauthor{cai2016neural} \shortcite{cai2016neural} use Gated Combination Neural Networks and LSTM to present both character sequences and partially segmented word sequences, combining word scores and link scores for segmentation. Our work is in line with their work in directly using word information for CWS. In contrast, our method is conceptually simpler by directly using word embeddings. In addition, our work aims at domain-adaptation, rather than training from scratch. 

\subsection{Cross-Domain CWS}
Supervised, semi-supervised and unsupervised approaches have been proposed for domain adaptation for CWS. \citeauthor{chen2017adversarial} \shortcite{chen2017adversarial} use an Adversarial Network to learn shared knowledge for different segmentation criteria and domains. This approach requires annotated data in the target domain. 

However, one challenge for cross-domain CWS is the lack of such annotated data. \citeauthor{liu2012unsupervised} \shortcite{liu2012unsupervised} propose an unsupervised model, in which they use features derived from character clustering, together with a self-training algorithm to jointly model CWS and POS-tagging. This approach is highly time-consuming \cite{qiu2015word}. Another challenge is the segmentation of domain-specific noun entities. In a task of segmenting Chinese novels, \citeauthor{qiu2015word} \shortcite{qiu2015word} design a double-propagation algorithm with complex feature templates to iteratively extract noun entities and their context, to improve segmentation performance. This approach still relies heavily on feature templates. Similarly, our model does not require any annotated target data. In contrast to their work, our model is efficient and feature-free.

\subsection{CWS Using Embeddings}
There are CWS models deploying embeddings. \citeauthor{ma2015accurate} \shortcite{ma2015accurate} and \citeauthor{deng2018improved} \shortcite{deng2018improved} propose embedding matching CWS models, in which embeddings of characters in a sequence are compared with high dimensional representations of CWS-specific actions (e.g., separation and combination) or CWS-specific labels (e.g., \emph{B/M/E/S}). Then each character is labeled according to the similarity between its embedding and the high dimensional representation. 

Particularly, \citeauthor{zhou2017word} \shortcite{zhou2017word} propose to use character embeddings trained on a word-based context to improve the performance of existing neural CWS models, which is similar to our approach in terms of making use of CWS-oriented word embeddings derived with automatically segmented raw corpus. However, in their work, when doing cross-domain CWS, word embeddings are fed into the baseline neural model trained on a large annotated general corpus, with annotated special domain data as the development set. In our model, on the contrary, word embeddings are used directly for CWS with a non-parametric decoder, which does not require to re-construct the baseline model, and annotation is not required at all for the special domain data. 


\section{Word-Embedding-Based CWS}

The overall architecture of our method is shown in Figure 1. Given a baseline segmenter and a target domain raw corpus $T$, we obtain an automatically segmented corpus $T^{'}$ by applying the baseline segmenter to the target corpus. We then execute our CWS-oriented word embedding model (Section 3.1) on $T^{'}$ to derive a set of word embeddings $E$. In addition, all tokens from $T^{'}$ are collected as a target domain dictionary $D$. Finally, $E$ and $D$ are used to re-segment $T$ with our word-embedding-based segmenter (Section \ref{3.2}).

\begin{figure}[t]
\centering
\includegraphics[width=2.8in]{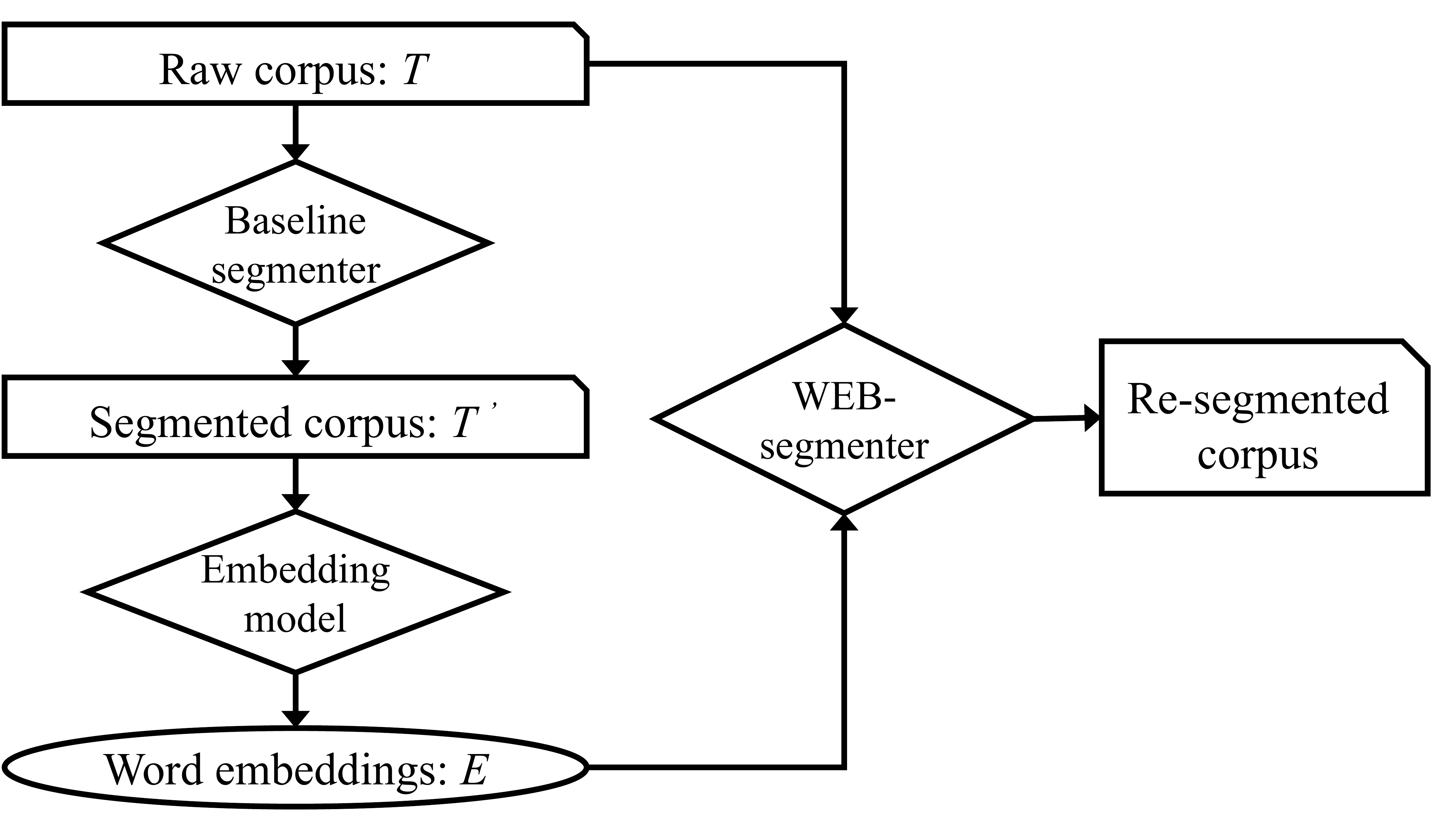}
\caption{The pipeline of WEB-CWS.}
\label{fig1}
\end{figure} 

\subsection{CWS-Oriented Word Embedding Model}
We use a CWS-oriented model modified from the Skip-gram model \cite{mikolov2013distributed} to derive word embeddings. A typical Skip-gram model using negative sampling tries to maximize the following objective \cite{mikolov2013distributed}:

\begin{equation}
\linespread{0.8}
\selectfont
\sum_{(w, c) \in P} \log{\sigma(v_w \cdot v_{c}^{\top})} + \sum_{(w, c^{'}) \in N} \log{\sigma(-v_w \cdot v_{c^{'}}^{\top})}
\end{equation}

\noindent with $P$ being the set of positive samples $(w, c)$ consisting of a target word $w$ and a context word $c$, $N$ being the set of negative samples $(w, c^{'})$ consisting of a target word $w$ and a word $c^{'}$ drawn randomly from a noise distribution $P_n(w)$, $v_w$ being the word embedding of $w$, and $\sigma$ being the sigmoid activation function. 

Subsampling is applied when choosing the target word $w$ to reduce training time and to improve the quality of embeddings of rare words \cite{mikolov2013distributed}. For a natural language, the frequency distribution of all words is expected to obey Zipf's law: a word's frequency is  inversely proportional to its rank in the frequency table \cite{newman2005power}. This highly biased distribution makes the training of the Skip-gram model inefficient, in that very frequent words can make a large portion of training samples, but their embeddings may not change much after being seen for a certain time \cite{mikolov2013distributed}. Therefore, a subsampling method is used by \citeauthor{mikolov2013distributed} \shortcite{mikolov2013distributed}, with the probability for a word $w$ being sampled as:

\begin{equation}
\linespread{0.8}
\selectfont
p_{sub}(w) = min(1, \sqrt {\frac{\epsilon}{f(w)}})
\end{equation}

\noindent where $\epsilon$ is an arbitrarily chosen threshold, and $f(w)$ is the frequency of $w$.

Since word embeddings in our model are used for segmentation, we cannot directly use the training objective in \citeauthor{mikolov2013distributed} \shortcite{mikolov2013distributed}, which is designed for language modeling. To make the training objective more consistent with the goal of CWS, we modify the negative sampling Skip-gram model in various ways, including adding CWS-oriented negative samples, changing the method for subsampling multi-character words, normalizing the dot product of embeddings, and smoothing the weights of positive and negative samples in training.

\subsubsection{Context Negative Sampling}
When training a typical Skip-gram model, a target word and a word within its context window are taken together as a positive sample \cite{mikolov2013distributed}. From the perspective of CWS, it can be perceived as teaching the model how to correctly segment a sequence. From this perspective, we develop a method to generate negative samples from a word's context (i.e., words within the context window), in order to tell the model what the incorrect segmentations of a sequence are.

Given a target word $w$ and its context $C$, and $SL$/$SR$ as the sequence of characters on the left/right of $w$ within $C$, the proposed context negative sampling method generates negative samples in the following way: for any substring $s^{'}$ of $SL$ and $SR$, if $s^{'}$ is in the dictionary $D$ but not in $C$, ($w$, $s^{'}$) will be generated as a negative sample.


\subsubsection{In-Word Negative Sampling}
Another way to generate negative samples concerning CWS is to split multi-character words and combine its substrings as negative samples. For instance, given a multi-character target word $w=c_1c_2c_3$, supposing that all its substrings are in $D$, the proposed in-word negative sampling method will then generate the following negative samples: $(c_1, c_2), (c_1, c_3), (c_2, c_3), (c_1c_2, c_3)$ and $(c_1, c_2c_3)$. By doing so, our model is expected to learn not to split those multi-character words when segmenting.

\subsubsection{Subsampling Multi-Character Words}
In the Chinese language, there are some frequent multi-character words consisting of substrings which are also very frequent words themselves. For example, the Chinese word `d\`{a}nsh\`{i} (but)' can be decomposed into two words `d\`{a}n (but)' and `sh\`{i} (be)'. Due to the nature of the Skip-gram model, embeddings of frequent words are relatively close to each other since they co-occur with other words more frequently. This nature makes our model inclined to split such multi-character words when segmenting. Although the in-word negative sampling method proposed above is expected to prevent our model from incorrectly splitting multi-character words, we still want our model to pay more attention to the segmentation of such words. As a result, for subsampling, we will not discard a multi-character word $w$ if:

\begin{equation}
\linespread{0.8}
\selectfont
p_{sub}(w) < \frac{\mu}{N}\sum_{w_{sub} \in S \cap D} p_{sub}(w_{sub})
\end{equation}

\noindent where $S$ is the set of substrings of $w$, and $N$ is the size of all $w_{sub} \in S \cap D$, which is smoothed by a threshold $\mu$ (which is empirically set to 0.5 in our model). By doing so, we can keep those multi-character words whose substrings are more frequent words themselves. Our model is thus expected to learn better how to segment such words through samples generated with in-word negative sampling.

\subsubsection{Dot Product Normalization}
In the original Skip-gram model, the dot product of embeddings of two words is directly used as the input for the sigmoid layer \cite{mikolov2013distributed}. To make word embeddings derived from the CWS-oriented word embedding model more consistent with the metric used for segmentation as described in Section \ref{cos}, we modify the training objective described in Equation (1) as follows:

\begin{equation}
\linespread{0.8}
\fontsize{9.5pt}{\baselineskip}
\selectfont
\sum_{(w, c) \in P} \log{\sigma(||v_w \cdot v_{c}^{\top}||)} + \sum_{(w, c^{'}) \in N} \log{\sigma(-||v_w \cdot v_{c^{'}}^{\top}||)}
\end{equation}

\subsubsection{Smoothing Class Weights}
For any target word, when training the CWS-oriented word embedding model, only one positive sample but many negative samples are generated. To balance the influence of positive and negative samples, a different weight is assigned to each class as follows:

\begin{equation}
\linespread{0.8}
\fontsize{9.5pt}{\baselineskip}
\selectfont
class\_weight=
\begin{cases}
1.0 & , positive\\
(\frac{N_{pos}}{N_{neg}} + \eta)/(1 + \eta) & , negative
\end{cases}
\end{equation}

\noindent where $N_{pos}$ and $N_{pos}$ are the amount of positive and negative samples respectively, and $\eta$ is a smooth factor. The smooth factor can prevent the weight of negative samples being too low when negative samples are much more than positive samples.

\subsection{Word-Embedding-Based Segmentater} \label{3.2}
In WEB-CWS, we formalize the process of segmenting a sequence as a problem of hypotheses-based Viterbi decoding \cite{forney1973viterbi}: given a sequence, generating segmentation hypotheses character-wise from the first to the last character, and then searching for the optimal path according to the predefined metric of probability.

\subsubsection{Hypothesis Generation}
Given a sentence consisting of $n$ characters $S=<\!c_0\!>c_1c_2...c_n<\!c_{n+1}\!>$ ($c_0$ and $c_{n+1}$ are markers of the beginning/end of a sentence), we generate segmentation hypotheses character-wise from $c_0$ to $c_{n+1}$. At each time step $t$, a hypothesis $h_t$ is defined as:

\begin{equation}
\linespread{0.8}
\selectfont
h_t=
\begin{cases}

SEG_t: [w_{0\{{c_{0}}\}}w_{1\{c_1...c_g\}}...w_{{m}\{c_{j}...c_k\}}] \\
BUF_t: [c_{k+1}...c_t] \\
M_t = m+1
\end{cases}
\end{equation}

\noindent which includes a partial segmentation container $SEG_t$, a buffer container $BUF_t$, and the number of segmented words $M_t$. In $h_t$, characters $c_0c_1...c_k$ are segmented into words $w_0w_1...w_m$ stored in $SEG_t$; characters $c_{k+1}..c_t$ remain unsegmented and are stored in $BUF_t$. For the initial hypothesis $h_0$ and the final hypothesis $h_{n+1}$, the buffer container will be empty.

Given a character $c_{t+1}$ and a hypothesis $h_t$, $h_{t+1}$ can be generated in two ways, by either appending $c_{t+1}$ to $BUF_t$ ($t \neq n$), or first moving the sequence in $BUF_t$ into $SEG_t$ as a new word, and then appending $c_{t+1}$ to $BUF_t$ ($t \neq n$). In the former case:

\begin{equation}
\linespread{0.8}
\selectfont
h_{t+1}=
\begin{cases}

SEG_t: [w_{0\{{c_{0}}\}}...w_{{m}\{c_{j}...c_k\}}] \\
BUF_t: [c_{k+1}...c_tc_{t+1}] \\
M_t = m+1
\end{cases}
\end{equation}

\noindent In the latter case:

\begin{equation}
\linespread{0.8}
\fontsize{9.5pt}{\baselineskip}
\selectfont
h_{t+1}=
\begin{cases}

SEG_t = w_{0\{{c_{0}}\}}...w_{{m}\{c_{j}...c_k\}}w_{{m+1}\{c_{k+1}...c_t\}} \\
BUF_t = c_{t+1}\\
M_t = m+2
\end{cases}
\end{equation}

Particularly, when generating the hypothesis for $c_{n+1}$ (the end of sentence marker), the sequence in $BUF_n$ has to be moved into $SEG_n$, and $c_{n+1}$ also needs to be moved into $SEG_n$ as a new word. 


If a word $w$ is not in the dictionary $D$, we cannot get its embedding. As a result, any hypothesis containing $w$ in the segmentation container will be discarded. Moreover, to reduce search space, once the size of the sequence in a buffer container reaches a threshold $m$ (i.e., the maximum word length), this sequence will be moved into the segmentation container when generating hypotheses at the next time step.

\subsubsection{Probability Calculation}\label{cos}
The log probability of a hypothesis $h_t$ is defined as:

\begin{equation}
\linespread{0.8}
\fontsize{9.5pt}{\baselineskip}
\selectfont
\log{p(h_t)} =
\begin{cases}
0, & t = 0 \\
\frac{1}{M_t-1}\log \prod_{i=1}^{M_t-1} p(w_i|SEG_t, f), & t \neq 0
\end{cases}
\end{equation}

\noindent where $f$ is the window size (e.g., if $f=2$, $p(w_i|SEG_t, f)$ will be decided by $w_{i-1}$ and $w_{i-2}$). 

In WEB-CWS, we use cosine similarity between embeddings of two words as the metric of probability. Given a hypothesis $h_t$, $p(w_i|SEG_t, f)$ is calculated as follows:

\begin{equation}
\linespread{0.8}
\fontsize{8.5pt}{\baselineskip}
\selectfont
p(w_i|SEG_t, f) =
\begin{cases}
e^0, & i=0\\
e^{\frac{1}{min(f, i)}\sum\limits_{j=1}^{min(f, i)}\cos(v_{w_i}, v_{w_{i-j}})}, & i \neq 0
\end{cases}
\end{equation}

\noindent where $\cos(v_{w_i}, v_{w_{i-j}})$ refers to the cosine similarity between embeddings of $w_i$ and $w_{i-j}$.

Given a hypothesis $h_t$, the log probability of $h_{t+1}$ can be dynamically computed as:

\begin{equation}
\linespread{0.8}
\fontsize{9.5pt}{\baselineskip}
\selectfont
\begin{split}
\log{p(h_{t+1})} &= \frac{1}{M_{t+1}-1}((M_t-1)\log{p(h_t)} \\
& + \sum_{w_i \in \{SEG_{t+1} - SEG_{t}\}}\log{p(w_i|SEG_{t+1}, f)})
\end{split}
\end{equation}

\subsubsection{Dynamic Beam-Size and Maximum Word Length}
Theoretically, a sequence of length $n$ can have at most $2^{n-1}$ possible segmentations. By discarding hypotheses containing out-of-vocabulary (OOV) words and setting the maximum word length, the search space can be significantly reduced. The very limited search space makes dynamically deciding the beam-size and the maximum word length possible.

Given an initial beam-size $k$, at each time step $t$, the segmenter will only keep at most top $k$ hypothesis sorted by log probabilities in a descending order. Some hypotheses will also be discarded due to OOV words and the maximum word length limit. As a result, it is sometimes possible for a sequence to have no hypothesis at all after some time steps. Once it happens, the segmenter will increase the beam-size by 10 and the maximum word length by 1, and then re-generate hypothesis from the beginning, till at least one hypothesis is generated at the final time step. 

Dynamic beam-size and maximum word length ensure that for each sequence, at least one segmentation (the one given by the baseline segmenter) will be generated as the final segmentation result. This mechanism can guarantee the efficiency and reliability of the segmenter at the same time. 

\subsubsection{Similarity Score Look-up}
To improve the decoding speed, we pre-calculate the cosine similarity of all word pairs co-occurring at least once in the automatically segmented corpus, and store them in a file (which only consists of millions of word pairs for a Chinese novel). In doing so, when decoding, for most word pairs, we only need to look up to this file for similarity scores, which can significantly improve the decoding speed. According to later experiments, this look-up strategy can cover about 92\% of the similarity calculation needed for decoding.

\section{Experiments}
We conduct experiments on various datasets in different domains to thoroughly evaluate the performance of our model.

\subsection{Experimental Setup} \label{4.1}

\subsubsection{Datasets}
We evaluate our model in terms of cross-domain CWS on five datasets, including three Chinese novel datasets \cite{qiu2015word}: DL (\emph{DouLuoDaLu}), FR (\emph{FanRenXiuXianZhuan}) and ZX (\emph{ZhuXian}), and two CWS datasets in special domains \cite{qiu2015construction}: DM (dermatology) and PT (patent). We use the standard split for all datasets as they are published. Raw test data is also included for deriving word embeddings.  

Statistics of these datasets are shown in Table \ref{tab2}. Since there are no gold segmentation of full novels for three Chinese novel datasets, their statistics are based on the segmentation given by the baseline segmenter.

\begin{table}[t]
\linespread{0.8}
\centering
\fontsize{8.8pt}{\baselineskip}
\selectfont
\begin{tabular}{@{}lcccccc@{}}
\toprule[1pt]
\multicolumn{1}{c}{Dataset}     & \multicolumn{2}{c}{Sentence (K)} & \multicolumn{2}{c}{Token (K)} & \multicolumn{2}{c}{Character (K)} \\ \midrule
 & \multicolumn{1}{c}{Full} &  \multicolumn{1}{c}{Eval} & \multicolumn{1}{c}{Full} &  \multicolumn{1}{c}{Eval} & \multicolumn{1}{c}{Full} &  \multicolumn{1}{c}{Eval} \\
DL  & 40 & 1                       & 1,982 & 32                      & 2,867 & 47                       \\ 
FR  & 148 & 1                      & 5,004 & 17                      & 7,126 & 25                       \\ 
ZX  & 59 & 1                       & 2,131 & 21                      & 3,006 & 31                       \\ 
DM  & 32 & 1                       & 709 & 17                        & 1,150 & 30                       \\ 
PT  & 17 & 1                       & 556 & 34                        & 903 & 57                         \\
\bottomrule[1pt]
\end{tabular}
\caption{Statistics of full and evaluation datasets.}
\label{tab2}
\end{table}

\subsubsection{Pre-Processing} 
In some studies, pre-processing is applied in order to improve the performance of CWS models, including substituting consecutive digits and English letters, Chinese idioms and long words with unique symbols \cite{cai2016neural,cai2017fast,chen2015long}. However, we do not deploy such techniques for fair comparison, focusing only on the possible improvements brought by word embeddings. The only pre-processing adopted in our model is to first split a sentence with a set of pre-defined delimiters: characters that are not Chinese characters, English letters or digits. Those fragments of a sentence are then fed into the segmenter, and a complete segmented sentence is returned by reassembling the segmented fragments and delimiters in the original order.

\subsubsection{Hyperparameters}
Hyperparameters used in our WEB-CWS model are explained and their values are displayed in Table \ref{tab3}. All hyperparameters are tuned on a small excerpt of ZX, which consists of 300 sentences \cite{qiu2015word}. It is worth noting that, according to \citeauthor{mikolov2013distributed} \shortcite{mikolov2013distributed}, for each positive sample, the optimal number of negative samples drawn from a noise distribution is usually between 5 to 20. However, in our model, we find that, for each target word, drawing one negative sample from a noise distribution is good enough, which may be caused by the large amount of negative samples generated by context and in-word negative sampling. Also, \citeauthor{mikolov2013distributed} \shortcite{mikolov2013distributed} report that the unigram distribution raised to the 3/4ths power is better than the uniform distribution for negative sampling. But in WEB-CWS, using the uniform distribution leads to better segmentation results.

\begin{table}[t]
\linespread{0.8}
\centering
\fontsize{6.5pt}{\baselineskip}
\selectfont
\begin{tabular}{p{0.1in}p{1.9in}c}
\toprule[1pt]
\multicolumn{1}{c}{H-param} & \multicolumn{1}{c}{Explanation} & \multicolumn{1}{c}{Value}\\ \midrule
$\epsilon$ & threshold for overall subsampling & $10^{-5}$ \\ 
$\mu$ & threshold for multi-character word subsampling & 0.5 \\ 
$P_n$ & noise distribution for general negative sampling & uniform \\ 
$n$ & number of general negative samples per word & 1 \\ 
$d$ & dimension of word embeddings & 100 \\ 
$\eta$ & smoothing factor for class weights & 0.2 \\ 
$f$ & window size & 4 \\ 
$m$ & initial maximum word length & 5 \\ 
$k$ & initial beam-size & 10 \\ 
$e$ & number of epochs in Skip-gram training & 1 \\ 
\bottomrule[1pt]
\end{tabular}
\caption{Hyperparameters used in WEB-CWS.}
\label{tab3}
\end{table}

\begin{table*}[t]
\linespread{0.8}
\centering
\fontsize{10.5pt}{\baselineskip}
\selectfont
\begin{tabular}{@{}lccccc@{}}
\toprule[1pt]
\multicolumn{1}{c}{Dataset} & \multicolumn{1}{c}{Baseline (\%)} & \multicolumn{1}{c}{WEB-CWS (\%)} & \multicolumn{1}{c}{IR\_WEB (\%)} & \multicolumn{1}{c}{IR\_Qiu\&Zhang (\%)} & \multicolumn{1}{c}{IR\_Liu\&Zhang (\%)} \\ \midrule
DL  & 90.5            & \bf{93.5}         & \bf{+3.3}    & +1.6   & +2.1    \\ 
FR  & 85.9            & \bf{89.6}         & \bf{+4.3}    & +2.9   & +2.5    \\ 
ZX  & 86.8            & \bf{89.6}         & \bf{+3.2}    & +2.2   & +2.4    \\ 
DM  & 77.9            & \bf{82.2}         & \bf{+5.5}    & -      & +1.8    \\ 
PT  & 84.6            & \bf{85.1}         & \bf{+0.6}    & -      & +0.2    \\
\bottomrule[1pt]
\end{tabular}
\caption{F-measures of the baseline segmenter and WEB-CWS on datasets in special domains, and F-measure improvement rates (IR) of WEB-CWS, \citeauthor{qiu2015word} \shortcite{qiu2015word} and \citeauthor{liu2012unsupervised} \shortcite{liu2012unsupervised}.}
\label{tab4}
\end{table*}

\subsubsection{Evaluation}
For consistency, all segmentation results are automatically calculated with the script provided in the SIGHAN Bakeoff \cite{emerson2005second} and are reported as word F-measures.

\subsection{Baseline Segmenter}
Two state-of-the-art CWS models trained on a People's Daily corpus in 2000 January are tested. One is a joint word segmentation and POS-tagging model \cite{zhang2010fast}, and the other is a word-based neural CWS model \cite{cai2017fast}. When training both models, default settings are used, except that the maximum word length in \citeauthor{cai2017fast}'s model is set to 5, which is in line with the setting of WEB-CWS.

On the evaluation set of PKU \cite{emerson2005second}, both models yield comparable results, but on the evaluation set of DL, \citeauthor{zhang2010fast}'s model (F-measure = 0.905) performs better than \citeauthor{cai2017fast}'s model (F-measure = 0.849). It is very possible that \citeauthor{zhang2010fast}'s model can handle cross-domain CWS more effectively. As a result, we choose \citeauthor{zhang2010fast}'s model as the baseline segmenter for following experiments.

\section{Results}

Results in Table \ref{tab4} show that our WEB-CWS model can obviously improve CWS on four datasets in special domains, including DL, FR, ZX and DM, with an increase of over 3.0\% in F-measure. Those four datasets are all in domains (novel and dermatology) which are very different from that of the baseline segmenter (newspaper). This result suggests that WEB-CWS can effectively improve cross-domain CWS.

However, on another dataset in the special domain PT, the improvement is not significant. There are two possible reasons for this result. First, the size of PT is the smallest among all datasets, which may make the quality of  word embeddings unsatisfactory. Second, the PT dataset contains a huge amount of decimal points (e.g., `3.14'), percentage signs (e.g., `28\%'), hyphens (e.g., `pMIV-Pnlp０') and very long English strings (e.g., `agctgagtcg'), which are all cases that cannot be handled by WEB-CWS without corresponding pre-processing techniques.


\subsection{Comparison with State-of-the-Art Models} We also compare WEB-CWS with two state-of-the-art semi-supervised and unsupervised cross-domain CWS models by \citeauthor{qiu2015word} \shortcite{qiu2015word} and \citeauthor{liu2012unsupervised} \shortcite{liu2012unsupervised}, both of which use the same baseline model proposed by \citeauthor{zhang2010fast} \shortcite{zhang2010fast} as used in our model. We adopt the method of combining character clustering and self-training in \citeauthor{liu2012unsupervised} \shortcite{liu2012unsupervised} with datasets in our experiments. Results of the model in \citeauthor{qiu2015word} \shortcite{qiu2015word} are directly copied from the corresponding paper. 

Results in Table \ref{tab4} show that WEB-CWS outperforms these two state-of-the-art models with a large margin in terms of F-measure. Particularly, on the DM dataset, \citeauthor{liu2012unsupervised}'s model only achieves a relatively low F-score improvement rate (1.8\%), which is likely to be caused by the large difference between the source and target domains. This result suggests that WEB-CWS is more robust to domain dissimilarity compared with self-training.

\subsection{Run Time for Decoding} 
We test the run time for our decoder on a 3.5 GHz Intel Core i7 CPU. On all five test sets, the average decoding speed is 20.3 tokens per millisecond, when the initial beam-size is set to 10. In the work of \citeauthor{zhou2017word} \shortcite{zhou2017word}, the decoding speed of their model is 14.7 tokens per millisecond for greedy segmentation. However, these results cannot be compared directly since they are produced on different machines. Our similarity look-up strategy is proved to be efficient in improving the decoding speed. 

\section{Analysis and Discussion}
In order to assess the effect of negative sampling and subsampling methods, we conduct a series of ablation experiments. A detailed analysis is presented to understand in what way WEB-CWS can improve cross-domain CWS. All experiments and analyses in this section are carried out on three datasets with most significant improvements in F-measure: DL, FR and DM.

\subsection{Ablation Experiments}
In ablation experiments, we study the influence of two CWS-oriented negative sampling and 
multi-character words subsampling. Results in Table \ref{tab6} show that WEB-CWS using word embeddings derived with the basic Skip-gram model (`\emph{basic}') performs obviously worse than the baseline segmenter. When CWS-oriented negative sampling is applied alone, either context (`\emph{c\_n}') or in-word (`\emph{w\_n}') negative sampling, the performance of WEB-CWS is obviously better than or similar to that of the baseline segmenter. When both CWS-oriented negative sampling methods are applied together (`\emph{c\_w\_n}'), WEB-CWS is ensured to obviously outperform the baseline segmenter. Also, when multi-character subsampling (`\emph{m\_s}') is applied, the performance of WEB-CWS can further improve a little.

\begin{table}[t]
\linespread{0.8}
\centering
\fontsize{9.5pt}{\baselineskip}
\selectfont
\begin{tabular}{p{1.1in}ccc}
\toprule[1pt]
\multicolumn{1}{c}{Model} & \multicolumn{1}{c}{DL (\%)} & \multicolumn{1}{c}{FR (\%)} & \multicolumn{1}{c}{DM (\%)}\\ \midrule
Baseline             & 90.5          & 85.9          & 77.9         \\ 
WEB-basic            & 76.2          & 72.2          & 66.6         \\ 
WEB + c\_n           & 92.7          & 89.3          & 75.8         \\ 
WEB + w\_n           & 93.0          & 89.5          & 76.2         \\ 
WEB + c\_w\_n        & 93.3          & 89.2          & 79.7         \\ 
WEB + c\_w\_n + m\_s & \bf{93.5}     & \bf{89.6}     & \bf{82.2}    \\ 
\bottomrule[1pt]
\end{tabular}
\caption{F-measures of WEB-CWS with different subsampling and negative sampling methods.}
\label{tab6}
\end{table}

\subsection{Improvements in Noun Entity Segmentation}
To see which words are incorrectly segmented by the baseline segmenter but correctly by WEB-CWS, all words occurring at least ten times in the three datasets are sorted in a descending order, by improvements in terms of segmentation precision. Table \ref{tab8} displays the ten most improved words. As shown in Table \ref{tab8}, among the ten most improved words, seven words are domain-specific noun entities, including person names, disease names and chemical compound names. For some noun entities (e.g., \emph{glucocorticoid}), even if the baseline segmenter can rarely segment them correctly, WEB-CWS can still find the correct segmentation in most cases. This result suggests that WEB-CWS is especially effective in segmenting domain-specific noun entities.

\begin{table}[t]
\centering
\includegraphics[width=2.6in]{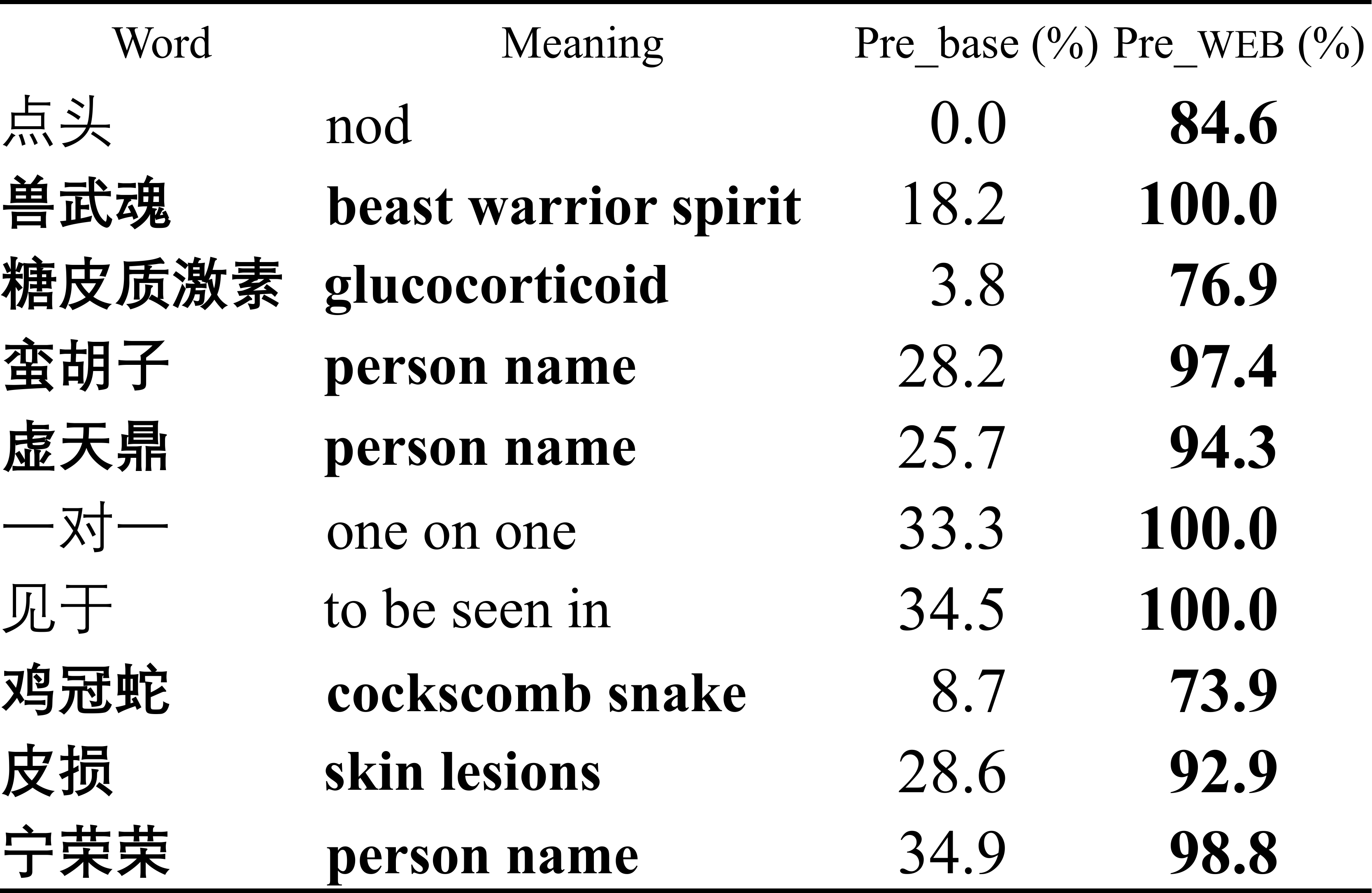}
\caption{Ten most improved words in terms of segmentation precision. Pre\_base: segmentation precision by the baseline segmenter. Pre\_WEB: segmentation precision by WEB-CWS.}
\label{tab8}
\end{table}

\section{Conclusion}
We have proposed WEB-CWS, a semi-supervised model that can be used to effectively improve cross-domain CWS. Our model only requires a baseline segmenter and a raw corpus in the target domain, deploying only word embeddings for CWS. WEB-CWS obviously improves the performance of the state-of-the-art baseline segmenter on four datasets in special domains, especially in segmenting domain-specific noun entities.

\section{Acknowledgment}
This paper was partially supported by the National Natural Science Foundation of China (No. 61572245). Thanks to You Wang et.al. and the anonymous reviewers for their constructive and insightful comments on this paper.

\bibliography{cws}
\bibliographystyle{acl_natbib}

\end{document}